\documentclass{article}

\usepackage[utf8]{inputenc}
\usepackage[T1]{fontenc}
\usepackage{hyperref}
\usepackage{url}
\usepackage{booktabs}
\usepackage{amsmath}
\usepackage{amssymb}
\usepackage{microtype}
\usepackage{natbib}
\usepackage{graphicx}
\usepackage{float}
\usepackage{placeins}
\usepackage[accepted]{mlsys2025}
\usepackage{fancyvrb}
\usepackage{float}
\usepackage{algorithm}
\usepackage{algorithmic}
\RecustomVerbatimEnvironment{verbatim}{Verbatim}{fontsize=\footnotesize,fontfamily=cmtt}
\usepackage{tikz}
\usetikzlibrary{arrows.meta,calc}
\usepackage{graphicx}

\mlsystitlerunning{Reproducible LoRA Inference Through Deterministic Split-K Reduction}

\hypersetup{
  pdftitle={Reproducible LoRA Inference Through Deterministic Split-K Reduction},
  pdfauthor={Dhayaneshwar Balusamy; Ankur Mohan},
  pdfsubject={arXiv Preprint},
  pdfkeywords={Split-K, LoRA, vLLM, deterministic inference}
}

\begin{document}

\twocolumn[{
\mlsystitle{Reproducible LoRA Inference Through Deterministic Split-K Reduction}

\begin{center}
\begin{minipage}[t]{0.46\textwidth}
\centering
\textbf{Dhayaneshwar Balusamy}\\[4pt]
JPMorgan Chase \& Co\\[4pt]
{\footnotesize
\href{mailto:dhayaneshwar.balusamy@chase.com}
{\textcolor{black}{\texttt{dhayaneshwar.balusamy@chase.com}}}
}
\end{minipage}
\hfill
\begin{minipage}[t]{0.46\textwidth}
\centering
\textbf{Ankur Mohan}\\[4pt]
JPMorgan Chase \& Co\\[4pt]
{\footnotesize
\href{mailto:ankur.mohan@chase.com}
{\textcolor{black}{\texttt{ankur.mohan@chase.com}}}
}
\end{minipage}
\end{center}

\vskip 0.3in

\begin{abstract}

Reproducibility is important for LoRA-adapted language
models used in repeated evaluation, auditing, and
decision-making. However, Split-K kernels used to introduce parallelism in low-rank LoRA multiplications can have the unintended consequence of numerical nondeterminism when partial
results are combined using floating-point atomic
additions. We investigate this non-determinism in vLLM across GSM8K mathematical reasoning, MBPP
code generation, and Jev-style structured toxicity decisions
on Civil Comments and introduce a deterministic Triton Split-K implementation. Under identical-input replay, Atomic
Split-K=8 produces different FP32 shrink outputs in 97.61\%
of comparisons, while our Deterministic Split-K=8
shows no observed variation. Repeated atomic inference changes
generated outputs in 17.30\% of GSM8K and 14.96\% of MBPP
comparisons, sometimes altering task correctness.
Structured-decision probabilities vary in 81.47\% of atomic
comparisons, with decision flips. Controlled tracing
demonstrates how shrink-output differences propagate into
the KV cache and how accumulated cache differences can
change token decisions and final answers. Our deterministic
Triton Split-K=8 implementation preserves parallelism,
achieving 16.8-21.1\% latency-based speedup over unsplit
reduction with only 1.02-2.62\% additional latency
relative to Atomic Split-K=8.

\end{abstract}

}]

\section{Introduction}

Low-Rank Adaptation (LoRA) has become a widely adopted approach for adapting large language models (LLMs) to downstream tasks without duplicating their pretrained weights~\citep{hu2022lora}. By introducing compact, task-specific parameters, LoRA enables efficient specialization and deployment of multiple models on shared inference infrastructure. Systems such as Punica and SLoRA demonstrate the
importance of specialized GPU execution and efficient
adapter serving~\citep{chen2024punica,MLSYS2024_906419cd}. While inference systems are typically optimized for throughput and latency, reproducibility is also important when model outputs must be audited, compared, or used to make decisions. Repeated inference with identical model parameters, inputs, and decoding configurations should ideally produce consistent results. However, numerical properties of GPU execution can undermine this expectation even under greedy decoding ~\citep{he2025nondeterminism}.

We encountered this issue while evaluating LoRA-adapted guardrail models that assess generated content against safety, policy, and application-specific requirements, including toxicity, prompt-injection, and adherence to the supplied context. During evaluation, we observed inconsistent numerical outputs across nominally identical inference executions. Such variability raises concerns for guardrail applications, where small changes in prediction scores may affect threshold-based decisions and inconsistent results can complicate auditing and repeated evaluation. 
This motivated a systems-level investigation into
whether LoRA inference could provide repeatable
predictions without sacrificing the parallelism
needed for efficient serving.

To study this behavior systematically, we use three publicly available datasets. We evaluate GSM8K mathematical reasoning and MBPP Python code generation to investigate numerical instability during autoregressive generation. In these tasks, a difference in an early token selection changes the context for subsequent predictions, potentially leading to a different final numerical answer or generated program. However, changes in generated outputs do not necessarily imply changes in task correctness. We extend our investigation to Jev-style structured decisions using vLLM's structured-decision interface ~\citep{vllm2026structured} with the \texttt{noul} (yes/no) question type, evaluating the stability of returned toxicity probabilities and binary decisions on Civil Comments. This setting allows us to examine numerical variation in prediction scores and its potential effect on decisions near the classification boundary.

Our investigation focuses on the LoRA \emph{shrink} operation in vLLM~\citep{kwon2023efficient}. For a linear layer with a LoRA adapter, the output is computed as
\begin{equation}
Y = XW^{\mathsf T} + s\,(XA^{\mathsf T})B^{\mathsf T},
\label{eq:lora}
\end{equation}
where $A\in\mathbb{R}^{r\times H}$ and $B\in\mathbb{R}^{H_{\mathrm{out}}\times r}$ are the low-rank projection matrices, $r$ is the adapter rank, and $s$ is the scaling factor. The shrink operation, $XA^{\mathsf T}$, reduces the input hidden dimension to the much smaller adapter rank. To improve GPU utilization, Split-K parallelism divides the reduction dimension across multiple thread blocks. In the atomic Split-K implementation examined in vLLM v0.29.0, independently computed partial results are accumulated using floating-point atomic additions. The order of these additions depends on GPU execution scheduling, and because floating-point addition is not associative ~\citep{goldberg1991every}, different accumulation orders can produce different intermediate values despite identical inputs and weights.

The consequences of this numerical variation depend on its propagation through subsequent model computations. Differences in FP32 shrink outputs may disappear during precision conversion, while others may survive and perturb downstream activations and logits. A change in logits does not necessarily affect the selected token, but sufficiently large perturbations relative to the decision margin can change the prediction. During autoregressive generation, such a change may propagate through subsequent tokens, producing divergent responses and potentially different task outcomes.

We investigate these effects by comparing three LoRA
shrink implementations: Atomic Split-K=8, unsplit
Split-K=1, and our Deterministic Split-K=8 implementation.
In our implementation, independent GPU thread blocks
compute FP32 partial results, while integer atomic counters
identify the last-finishing block, which combines the
partials in a fixed order. We evaluate numerical
reproducibility through identical-input kernel replays,
its impact on model predictions and task correctness
through repeated inference, and the performance trade-offs
among the three implementations.

\subsection{Our Contributions}
\label{sec:contributions}

This work makes the following contributions:

\begin{enumerate}
    \item \textbf{Characterization of LoRA shrink nondeterminism.}
    We isolate numerical variability within vLLM's
    atomic Split-K LoRA shrink operation by replaying
    identical activations and adapter weights. Across
    prefill and decode, we quantify how unordered FP32
    accumulation produces different outputs under
    otherwise identical execution conditions.

    \item \textbf{Analysis of application-level consequences.}
    We systematically evaluate how numerical variability
    affects generated outputs and task correctness on
    GSM8K and MBPP, as well as toxicity probabilities and
    binary decisions through Jev-style structured-decision
    interface on Civil Comments. A detailed case study further traces numerical differences through the LoRA shrink operation and KV cache, using
controlled cache replacement to reproduce a changed token
decision and final answer.

\item \textbf{Deterministic parallel LoRA shrink.}
We implement a deterministic Triton Split-K kernel
that preserves parallelism and uses per-tile
counters to enable fixed-order FP32 reduction without
a separate reduction launch. Our implementation achieves
lower latency than an existing deterministic Split-K
kernel while eliminating observed run-to-run variability
under fixed execution configurations. It delivers an
average 19.2\% speedup over Split-K=1 with only 1.69\%
latency overhead relative to nondeterministic Atomic
Split-K=8.

\end{enumerate}

\section{Background: Split-K in LoRA Shrink} 
\label{sec:background}

The LoRA shrink operation in Equation~\ref{eq:lora} projects an input
$X\in\mathbb{R}^{M\times H}$ into a low-rank representation of shape
$[M,r]$. Unlike conventional transformer projections, whose output
dimensions are typically large, the shrink operation produces only
$r$ output features per token, where $r\ll H$. This makes the operation
a narrow matrix multiplication with limited output parallelism ~\citep{chen2024punica}.

For a GPU kernel that computes output tiles of size $B_M\times B_N$,
the number of independent thread blocks is approximately
\begin{equation}
N_{\mathrm{tiles}} =
\left\lceil\frac{M}{B_M}\right\rceil
\left\lceil\frac{r}{B_N}\right\rceil.
\label{eq:output_tiles}
\end{equation}
For example, with \(M=256\), \(r=16\), \(B_M=32\), and \(B_N=16\), the shrink operation exposes eight output tiles, corresponding to eight thread blocks in the unsplit computation. A conventional
transformer projection with a much wider output dimension can expose
hundreds or thousands of tiles for the same number of input tokens.
Consequently, even during prefill, a low-rank shrink operation may
not provide enough independent parallel work to utilize the GPU effectively.

The limitation can be more pronounced during autoregressive decode,
where the number of active tokens may be much smaller. Thus, both
single-token classification workloads, which primarily exercise
prefill, and longer generation workloads, which include repeated
decode steps, can benefit from additional parallelism along the
reduction dimension.

\subsection{Split-K Decomposition}

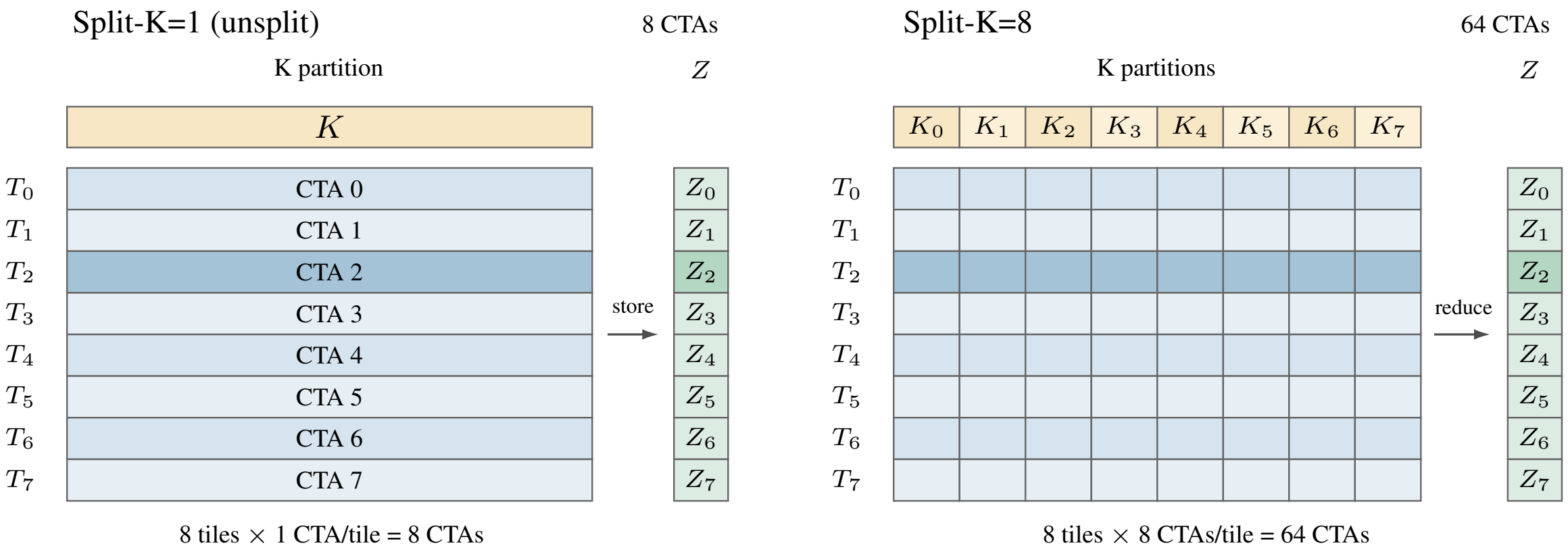
\begin{figure*}[t]
\centering

\definecolor{skgoldA}{RGB}{247,231,197}
\definecolor{skgoldB}{RGB}{252,241,216}
\definecolor{skblueA}{RGB}{214,228,239}
\definecolor{skblueB}{RGB}{231,239,245}
\definecolor{skblueFocus}{RGB}{165,195,215}
\definecolor{skgreenA}{RGB}{221,236,226}
\definecolor{skgreenFocus}{RGB}{180,215,195}

\begin{tikzpicture}[
  x=1cm, y=1cm,
  font=\footnotesize\rmfamily,
  cell/.style={
    draw=black!60,
    line width=0.45pt
  },
  arr/.style={
    draw=black!70,
    line width=0.6pt,
    -{Latex[length=1.6mm,width=1.05mm]}
  }
]

\path[use as bounding box]
  (0,-0.52) rectangle (15.85,5.10);


\node[anchor=west,font=\small\rmfamily]
  at (0.86,4.86) {Split-K=1 (unsplit)};

\node[anchor=east,font=\scriptsize]
  at (7.43,4.87) {8 CTAs};

\node[anchor=west,font=\small\rmfamily]
  at (8.99,4.86) {Split-K=8};

\node[anchor=east,font=\scriptsize]
  at (15.57,4.87) {64 CTAs};

\node[font=\scriptsize]
  at (3.49,4.42) {K partition};

\node[font=\scriptsize]
  at (11.59,4.42) {K partitions};

\node[font=\scriptsize]
  at (7.13,4.42) {$Z$};

\node[font=\scriptsize]
  at (15.24,4.42) {$Z$};


\filldraw[cell,fill=skgoldA]
  (0.93,3.66) rectangle (6.07,4.06);

\node at (3.50,3.86) {$K$};


\foreach \s in {0,...,7} {

  \pgfmathsetmacro{\xl}{9.02+0.645*\s}
  \pgfmathsetmacro{\xr}{\xl+0.645}

  \pgfmathtruncatemacro{\parity}{mod(\s,2)}

  \ifnum\parity=0
    \def\kshade{skgoldA}
  \else
    \def\kshade{skgoldB}
  \fi

  \filldraw[cell,fill=\kshade]
    (\xl,3.66) rectangle (\xr,4.06);

  \node[font=\scriptsize]
    at ({(\xl+\xr)/2},3.86)
    {$K_{\s}$};
}


\foreach \i in {0,...,7} {

  \pgfmathsetmacro{\ytop}{3.46-0.4075*\i}
  \pgfmathsetmacro{\ybot}{\ytop-0.4075}
  \pgfmathsetmacro{\yc}{(\ytop+\ybot)/2}

  \pgfmathtruncatemacro{\parity}{mod(\i,2)}

  \ifnum\i=2
    \def\shade{skblueFocus}
    \def\zshade{skgreenFocus}
  \else
    \ifnum\parity=0
      \def\shade{skblueA}
    \else
      \def\shade{skblueB}
    \fi
    \def\zshade{skgreenA}
  \fi

  \node[anchor=east,font=\scriptsize]
    at (0.75,\yc) {$T_{\i}$};

  \node[anchor=east,font=\scriptsize]
    at (8.84,\yc) {$T_{\i}$};


  \filldraw[cell,fill=\shade]
    (0.93,\ybot) rectangle (6.07,\ytop);

  \node[font=\scriptsize]
    at (3.50,\yc) {CTA \i};


  \foreach \s in {0,...,7} {

    \pgfmathsetmacro{\xl}{9.02+0.645*\s}

    \filldraw[cell,fill=\shade]
      (\xl,\ybot)
      rectangle ++(0.645,0.4075);
  }


  \filldraw[cell,fill=\zshade]
    (6.87,\ybot) rectangle (7.40,\ytop);

  \filldraw[cell,fill=\zshade]
    (15.03,\ybot) rectangle (15.56,\ytop);

  \node[font=\scriptsize]
    at (7.135,\yc) {$Z_{\i}$};

  \node[font=\scriptsize]
    at (15.295,\yc) {$Z_{\i}$};
}


\draw[arr]
  (6.22,1.83) -- (6.72,1.83);

\node[font=\tiny]
  at (6.47,2.10) {store};

\draw[arr]
  (14.31,1.83) -- (14.86,1.83);

\node[font=\tiny]
  at (14.60,2.10) {reduce};


\node[font=\scriptsize]
  at (3.52,-0.14)
  {8 tiles $\times$ 1 CTA/tile = 8 CTAs};

\node[font=\scriptsize]
  at (12.08,-0.14)
  {8 tiles $\times$ 8 CTAs/tile = 64 CTAs};

\end{tikzpicture}

\caption{
Thread-block parallelism in LoRA shrink.
Split-K=1 assigns one CTA per output tile,
whereas Split-K=8 assigns eight CTAs per tile
across independent K partitions. The highlighted
row illustrates one output tile.
}

\label{fig:splitk_parallelism}
\end{figure*}

Split-K increases parallelism by partitioning the reduction dimension
$H$ into $S$ subsets
$\mathcal{K}_1,\ldots,\mathcal{K}_S$.
Each partition computes a partial result:
\begin{equation}
Z_{ij}^{(s)} =
\sum_{k\in\mathcal{K}_s} X_{ik}A_{jk},
\qquad
Z_{ij} = \sum_{s=1}^{S}Z_{ij}^{(s)}.
\label{eq:partial}
\end{equation}

In the unsplit configuration, one thread block computes
each output tile. Split-K instead assigns $S$ independent
thread blocks per output tile, each processing a different
subset of the reduction dimension. This increases the
number of thread blocks from $N_{\mathrm{tiles}}$ to
$S N_{\mathrm{tiles}}$. As illustrated in
Figure~\ref{fig:splitk_parallelism}, for eight output
tiles, Split-K=8 increases the number of thread blocks
from 8 to 64.

This additional parallelism comes at the cost of storing or combining
partial results. Its benefit depends on the input dimensions, LoRA
rank, reduction overhead, and GPU architecture. When the original
output grid already provides sufficient parallelism, splitting the
reduction dimension may provide little benefit.

To illustrate this benefit, we profile the native
LoRA shrink operation on an NVIDIA RTX PRO 6000
Blackwell GPU (188 SMs). For a one-token decode
operation with BF16 operands, hidden dimension 3,072,
rank 16, and three fused QKV slices, Split-K=8
increases the number of thread blocks from 3 to 24 (8x).
Nsight Compute reports an increase in SM activity from 1.46\% to 8.06\%. Separately,
unprofiled CUDA-graph measurements show that complete
shrink latency, including output initialization,
decreases from 14.16 to 4.08~$\mu$s, a 3.47$\times$
speedup over Split-K=1.

\subsection{Order Dependence and Downstream Observability}

In the atomic Split-K implementation examined in this work,
independent thread blocks accumulate their FP32 partial results
through floating-point atomic additions. Although atomic operations ensure that concurrent updates do not overwrite one another, the order in which thread blocks accumulate their partial results can vary across executions.
Because floating-point addition is not associative
in general \cite{goldberg1991every, higham1993accuracy}, different
evaluation orders can produce different results:
\begin{equation}
\operatorname{fl}(\operatorname{fl}(a+b)+c)
\neq
\operatorname{fl}(a+\operatorname{fl}(b+c)).
\end{equation}

For example, consider three partial values
$a=10^{30}$, $b=-10^{30}$, and $c=1$. In finite-precision
arithmetic, evaluating $(a+b)+c$ produces $1$, whereas
$a+(b+c)$ produces $0$, because the addition of $1$ to
$-10^{30}$ is rounded away. Thus, different accumulation
orders can produce different results even when the operands
are identical.

The significance of these differences depends on subsequent
computation. Some FP32 differences disappear when converted to
a lower precision, while others survive and may perturb downstream
activations or logits. Such perturbations do not necessarily change
the selected token or task outcome. We therefore distinguish
numerical variation at the shrink operation from differences
that propagate to model predictions.

Accumulating Split-K partial results in a fixed order removes
schedule-dependent floating-point accumulation from this
operation \cite{demmel2013fast}. However, it does not guarantee equivalence to an
unsplit reduction or reproducibility of the entire model.

\section{Models, Datasets, and Adapters}
\label{sec:data}

The work originated from a numerical reproducibility
issue observed in an internal production system,
where the resulting deterministic kernel was
subsequently deployed. To characterize the issue
and share reproducible evidence without relying
on proprietary workloads, the experiments reported
here use publicly available models and datasets.

\subsection{Datasets and Evaluation}

We evaluate mathematical reasoning using
GSM8K~\citep{cobbe2021trainingverifierssolvemath}, which contains
7,473 training and 1,319 test problems. We reserve
5\% of the training set for validation using a fixed
seed, yielding 7,099 training and 374 validation
examples. Preprocessing removes calculator annotations
and replaces the original \texttt{\#\#\#\#} delimiter
with \texttt{Final answer:}. At evaluation, the model
receives the question and an instruction to provide
concise reasoning followed by a numerical answer.
We extract the last \texttt{Final answer:} value
and compare it with the reference using exact
numerical equality; missing or unparseable answers
are counted as incorrect.\footnote{\href{https://huggingface.co/datasets/openai/gsm8k}{GSM8K dataset}}

For code generation, we use the MBPP
\texttt{full} configuration~\citep{austin2021programsynthesislargelanguage},
containing 374 training, 90 validation, and 500 test
tasks. Prompts include the task description, optional
setup code, and public test assertions, but not the
reference implementation. Generated programs are
evaluated against the provided assertions in a
network-disabled, resource-limited Docker
container.\footnote{\href{https://github.com/google-research/google-research/tree/master/mbpp}{MBPP dataset}}

For toxicity classification, we use Civil
Comments~\citep{borkan2019nuanced}. Binary labels
are obtained by thresholding dataset toxicity scores
at 0.5. After removing empty comments and exact
duplicates across partitions, fixed-seed,
class-proportional sampling selects 20,000 training,
2,000 validation, and 1,000 test examples. The test
set contains 920 non-toxic and 80 toxic comments.
Each comment is evaluated through vLLM's
\texttt{noul} interface using the question
``Is this comment toxic?'' The returned probability
determines the predicted class, allowing us to
measure probability and decision stability.
\footnote{\href{https://huggingface.co/datasets/google/civil_comments}{Civil Comments dataset}}

\subsection{Model families and adapter configurations}

We evaluate Llama-3.2-3B-Instruct ~\citep{meta2024llama}, Gemma-4-E4B-it ~\citep{team2026gemma}, and NVIDIA-Nemotron-3-Nano-4B-BF16 ~\citep{blakeman2025nemotron} on GSM8K and MBPP. These models span different architectures: Llama uses a conventional dense Transformer, Gemma incorporates local and global attention, and Nemotron combines Mamba-2 and Transformer layers. This diversity allows us to examine LoRA shrink reproducibility across different model designs. For each model and task, we use separate LoRA adapters with ranks 16 and 32 and scaling factor $\alpha=2r$. For Civil Comments classification, we additionally evaluate Qwen3-0.6B ~\citep{yang2025qwen3} with a rank-16 LoRA adapter ($\alpha=32$), targeting attention and feed-forward projections.

All experiments use BF16 base-model and adapter weights
during inference, without quantization. Thinking mode
is disabled for Gemma and Nemotron. For each model,
task, and adapter configuration, we keep model weights,
input prompts, chat templates, and decoding settings
fixed across reduction schemes. We also verify that
the intended LoRA shrink kernel is executed.

\section{Execution Environment and Reduction Variants}
\label{sec:methods}

The GSM8K, MBPP, and Civil Comments experiments use BF16
model and adapter weights, FP32 shrink accumulation, and
no quantization. Inference uses tensor parallelism one and
a GPU-memory-utilization target of 0.85. GSM8K and MBPP
use a maximum model length and batched-token limit of
8,192, while Civil Comments uses 4,096 for both.

Prefix caching, asynchronous scheduling, and speculative
decoding are disabled. We evaluate sequential inference with
one active request at a time, along with fixed-batch inference
at batch sizes 1, 16, 32, and 64.
CUDA graphs are enabled for the main generation and timing
experiments, while identical-input shrink capture uses eager
execution. The GPU, software versions, adapter hashes, prompt
IDs, and kernel dispatch are recorded. We disable
\texttt{VLLM\_BATCH\_INVARIANT} to isolate the LoRA shrink
reduction, since enabling it can also change other inference
kernels.

The default LoRA shrink configuration in vLLM v0.29.0
selects Split-K=64 for $M<128$ and Split-K=8 for
$M\geq128$, where $M$ denotes the number of input
token rows. With batch-invariant execution enabled,
the examined shrink path instead selects Split-K=1.

We compare deterministic Split-K configurations with splits
$S\in\{2,4,8,16\}$. In our LoRA shrink
kernel benchmarks, Split-K=8 is 48.1\% faster than
Split-K=2 and 12.6\% faster than Split-K=4, while
being only 2.0\% slower than Split-K=16. We select
Split-K=8 for our main experiments as it offers
comparable performance to Split-K=16 with fewer
partitions, while also
aligning with vLLM's existing eight-way Split-K
configuration.

\begin{figure*}[t]
\centering
\begin{tikzpicture}[
  x=1cm,y=1cm,
  font=\footnotesize\rmfamily,
  line/.style={draw=black!72,line width=0.52pt},
  data/.style={
    draw=black!72,
    line width=0.62pt,
    -{Latex[length=1.4mm,width=0.95mm]}
  },
  control/.style={
    draw=black!48,
    dashed,
    line width=0.48pt
  },
  controlarrow/.style={
    control,
    -{Latex[length=1.4mm,width=0.95mm]}
  },
  kcell/.style={
    draw=black!65,
    line width=0.36pt
  },
  ctabox/.style={
    draw=black!68,
    fill=black!4,
    line width=0.50pt,
    minimum width=0.76cm,
    minimum height=0.39cm,
    inner sep=0pt
  },
  scratch/.style={
    draw=black!63,
    fill=blue!11,
    line width=0.48pt,
    minimum width=0.76cm,
    minimum height=0.48cm,
    inner sep=0pt
  },
  result/.style={
    draw=black!68,
    line width=0.52pt
  }
]

\path[use as bounding box]
    (0,-1.76) rectangle (16.15,3.93);

\node[font=\footnotesize] at (7.08,3.73)
    {One output tile, $S=8$};


\node at (1.02,3.04) {$X_{\mathrm{tile}}$};

\foreach \s in {0,...,7} {
  \pgfmathtruncatemacro{\shade}{mod(\s,2)}

  \foreach \j in {0,...,3} {
    \ifnum\shade=0
      \def\tone{green!16}
    \else
      \def\tone{green!7}
    \fi

    \draw[kcell,fill=\tone]
      (0.19+0.205*\s,1.77+0.22*\j)
      rectangle ++(0.205,0.22);
  }
}

\node[font=\scriptsize] at (1.01,1.54)
    {$B_M\!\times\!H$};

\node[font=\large] at (2.14,2.20)
    {$\times$};

\node at (2.69,3.04) {$A^{\mathsf T}$};

\foreach \s in {0,...,7} {
  \pgfmathtruncatemacro{\shade}{mod(\s,2)}

  \foreach \j in {0,1} {
    \ifnum\shade=0
      \def\tone{orange!21}
    \else
      \def\tone{orange!8}
    \fi

    \draw[kcell,fill=\tone]
      (2.43+0.27*\j,1.70+0.145*\s)
      rectangle ++(0.27,0.145);
  }
}

\node[font=\scriptsize] at (2.70,1.48)
    {$H\!\times\!r$};

\draw[data] (3.20,2.18) -- (3.64,2.18);


\node[font=\scriptsize,anchor=west]
    at (3.70,3.45)
    {eight independent CTAs};

\foreach \s in {0,...,7} {

  \pgfmathsetmacro{\xx}{4.09+0.91*\s}

  \node[ctabox] (cta\s) at (\xx,2.37)
    {\scriptsize CTA \s};

  \node[scratch] (p\s) at (\xx,1.18)
    {$P_{\s}$};

  \draw[data]
    (cta\s.south) -- (p\s.north);

  \draw[control]
    (cta\s.north) -- (\xx,3.07);

  \draw[
    line,
    draw=blue!50!black,
    line width=0.43pt
  ]
    (p\s.south) -- (\xx,0.66);
}

\node[font=\scriptsize,anchor=west]
    at (3.70,0.28)
    {FP32 scratch};


\draw[control]
    (4.09,3.07) -- (12.96,3.07);

\draw[controlarrow]
    (12.96,3.07) -- (12.96,2.82);

\draw[result,fill=black!2]
    (11.62,1.94) rectangle (14.65,2.82);

\node[font=\scriptsize] at (13.135,2.61)
    {completion counter $c$};

\foreach \i in {0,...,7} {
  \draw[
    draw=black!55,
    line width=0.36pt,
    fill=black!13
  ]
    (12.05+0.275*\i,2.16)
    rectangle ++(0.205,0.20);
}

\node[font=\scriptsize,anchor=west]
    at (14.74,2.25) {$c=8$};


\draw[
  draw=blue!50!black,
  line width=0.56pt
]
    (4.09,0.66) -- (10.46,0.66);

\draw[data,draw=blue!55!black]
    (7.28,0.66) -- (7.28,-0.36);


\draw[result,fill=blue!4]
    (4.81,-1.44) rectangle (11.50,-0.36);

\node[font=\scriptsize] at (8.16,-0.64)
    {last-finishing CTA: fixed-order FP32 sum};

\node[font=\normalsize] at (8.16,-1.09)
    {$P_0+P_1+\cdots+P_7$};

\draw[controlarrow]
    (13.17,1.94)
    -- (13.17,0.01)
    -- (11.15,0.01)
    -- (11.15,-0.36);

\node[font=\scriptsize,anchor=east]
    at (13.08,0.18)
    {$\mathrm{prev}=7$};


\draw[data]
    (11.50,-0.91) -- (14.58,-0.91);

\node[
  font=\scriptsize,
  fill=white,
  inner sep=1.5pt
]
    at (12.95,-0.68)
    {$\times\,\alpha/r$};

\foreach \j in {0,...,2} {
  \foreach \i in {0,...,2} {
    \draw[kcell,fill=blue!17]
      (14.69+0.24*\i,-1.27+0.24*\j)
      rectangle ++(0.24,0.24);
  }
}

\node at (15.05,-1.53)
    {$Z_{\mathrm{tile}}$};

\end{tikzpicture}

\caption{
Deterministic Split-K=8 LoRA shrink. Eight CTAs
compute FP32 partials for one output tile.
The last-finishing CTA combines them in a fixed
order and writes the scaled result.
}

\label{fig:deterministic_splitk}
\end{figure*}
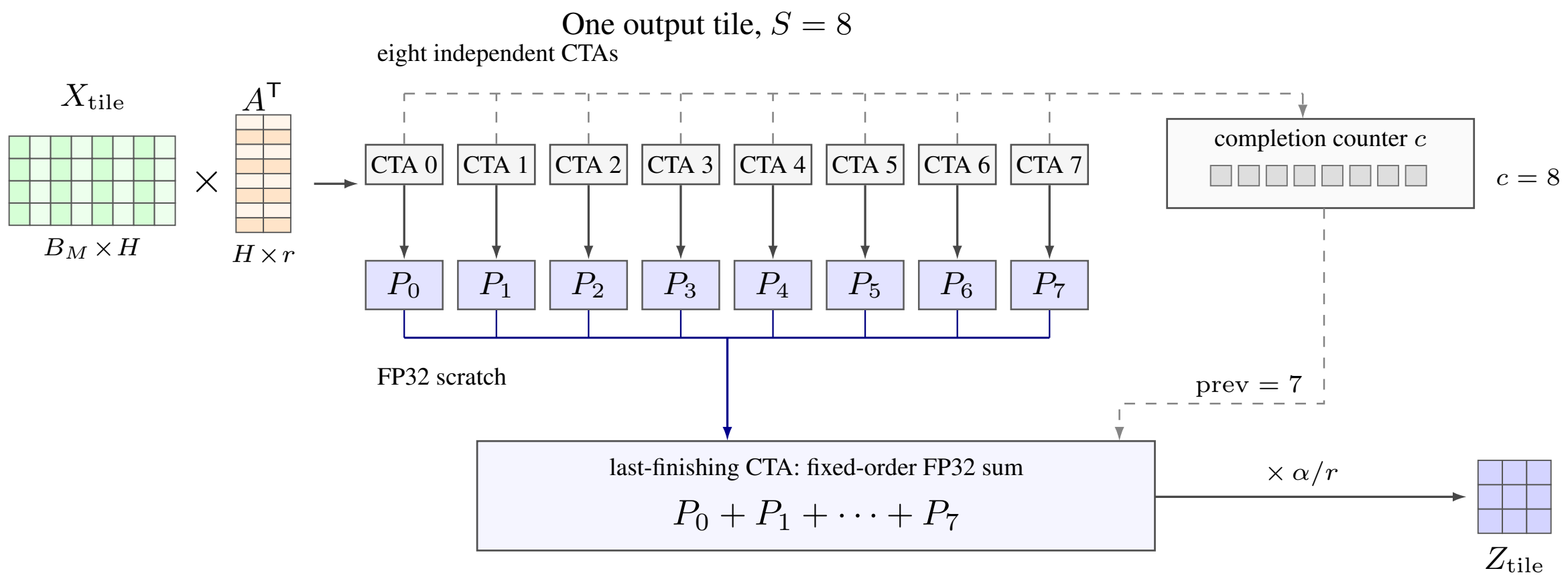

\subsection{Atomic Split-K=8}

In the atomic Split-K=8 baseline, the reduction dimension
is divided into eight partitions, each independently
computing partial output tiles. For each output tile,
the eight corresponding partial results are accumulated
using FP32 atomic additions. Since the order of these
additions depends on GPU scheduling, the accumulated values
may differ across executions. We denote this configuration
as \texttt{atomic\_sk8}.

\subsection{Unsplit Reduction (Split-K=1)}

In the unsplit Split-K=1 baseline, each GPU thread block computes
an output tile over the entire reduction dimension, without combining
partial results from multiple blocks. This eliminates cross-partition
accumulation within the LoRA shrink operation. However, its numerical
results may differ from Deterministic Split-K=8 because the two
implementations use different floating-point accumulation orders.

\subsection{Deterministic Split-K=8}
\label{sec:lastfinisher}

Our Triton~\citep{tillet2019triton} implementation preserves eight-way Split-K parallelism while replacing floating-point atomic accumulation with fixed-order summation. The kernel supports BF16 inputs and weights with FP32 shrink outputs. An initialization kernel clears the output tensor and per-tile integer completion counters. In the subsequent kernel, eight GPU thread blocks independently compute FP32 partial results for each output tile and store them in separate locations of a scratch buffer.

Before signaling completion, a GPU-wide memory
fence and a block-wide barrier ensure that the partial
result is published to memory. Each block then increments the per-tile
completion counter using a GPU-scoped
acquire-release integer atomic operation. The block observing the previous counter value of seven
is the last to signal completion~\citep{nvidia2025cuda}. It reads the eight
stored partial results, accumulates them in increasing
partition order using FP32 arithmetic, applies the
LoRA scaling factor, and writes the final output tile. Algorithm~\ref{alg:det_splitk} summarizes the
completion-counter synchronization and fixed-order
FP32 accumulation.

\begin{algorithm}[!htbp]
\caption{Deterministic Split-K=8 LoRA Shrink}
\label{alg:det_splitk}
\begin{algorithmic}[1]
\STATE Initialize output and completion counters to zero.

\FORALL{output tiles $t$ and splits $s \in \{0,\ldots,7\}$ in parallel}
    \STATE $p \gets \operatorname{compute\_fp32\_partial}(t,s)$
    \STATE $\mathrm{scratch}[t,s] \gets p$
    
    \STATE $\operatorname{gpu\_memory\_fence}()$
    \STATE $\operatorname{block\_barrier}()$

    \STATE $v \gets \operatorname{atomic\_increment}(\mathrm{counter}[t])$

    \IF{$v = 7$}
        \STATE $total \gets \mathrm{FP32}(0)$
        \FOR{$j=0$ {\bfseries to} $7$}
            \STATE $total \gets total + \mathrm{scratch}[t,j]$
        \ENDFOR
        \STATE $\mathrm{output}[t] \gets total \times scale$
        \STATE $\mathrm{counter}[t] \gets 0$
    \ENDIF
\ENDFOR
\end{algorithmic}
\end{algorithm}

Although the order in which blocks complete may vary, the
floating-point accumulation always follows the same partition
order. Integer atomic operations are used only to coordinate
completion, not to accumulate the partial results. The
implementation thus removes schedule-dependent floating-point
accumulation from LoRA shrink while retaining Split-K parallelism. 

We also explored an earlier deterministic implementation
that computed Split-K partials in parallel and combined
them using a separate kernel with a fixed pairwise
reduction tree. Although this provided a deterministic
accumulation order, the additional reduction launch
introduced overhead, particularly for small decode
workloads. This motivated our current design,
which integrates final accumulation into a
partial-computation kernel.

\section{Benchmark Protocol and Measurements}
\label{sec:eval}

We compare Atomic Split-K=8, Split-K=1, and
Deterministic Split-K=8 on a single NVIDIA RTX
PRO 6000 Blackwell GPU. All experiments use
BF16 model and adapter weights without quantization.
Reproducibility and performance are measured separately
to exclude diagnostic overhead from timing results.

\subsection{Generation Reproducibility}

For GSM8K and MBPP, we select 50 test prompts per
dataset using a fixed seed. The same prompts are
evaluated across three models, two LoRA ranks,
and three reduction methods. Each prompt is
generated ten times sequentially within the
same engine using greedy decoding, with a
4,096-token output limit.

Repetitions 1-9 are compared against repetition 0,
yielding 450 comparisons per configuration.
We measure differences in generated token sequences,
selected-token raw logits, and task correctness.
Logit differences are measured only at matching-prefix
positions with identical selected tokens, excluding
the first divergent token and subsequent positions.

\subsection{Identical-Input Shrink Replay}

To isolate kernel-level variability, we capture
shrink inputs, adapter weights, scaling factors,
and mapping information from ten prompts per
dataset under Split-K=1 in eager mode, covering
prefill and the first decode step.

Each captured operation is replayed using all
three reduction methods on identical inputs.
After warmup, five executions are collected per
method, with executions 1-4 compared against
execution 0. We record the fraction of repeat comparisons
with differing FP32 shrink outputs, the fraction
of differing FP32 output elements, and the mean
and maximum absolute differences.

\subsection{Inference Performance and Batch Sensitivity}

Performance is measured using 50 sequential
requests with exactly 150 generated tokens
per request and EOS stopping disabled,
across five fresh-engine trials per configuration.
We measure end-to-end latency, time to first token,
time per output token, and output throughput.
CUDA graphs are enabled, with initialization
and warmup excluded from measurements.

For batch-size experiments, we use
Llama-3.2-3B-Instruct on GSM8K with ranks 16
and 32, processing the same 64 prompts at batch
sizes 1, 16, 32, and 64. Timing uses three
fresh-engine trials. Separately, each fixed-batch
workload is repeated ten times to measure
within-batch reproducibility and differences
relative to batch-size-one execution.

\subsection{Structured Decision Models}

Recent structured decision models provide an alternative
to open-ended autoregressive generation when applications
require predefined decisions and associated probabilities.
Jev, introduced by TypeSafe AI, popularized this interface
through three decision primitives: selecting among
alternatives (\texttt{choice}), assigning an ordinal score
(\texttt{score}), and estimating the probability of a
binary yes/no decision (\texttt{noul})
\cite{almeida2026jev,li2026fast}. This approach is
particularly relevant to applications such as routing,
policy enforcement, and content moderation, where
structured decisions are more useful than generated text.

vLLM has recently incorporated support for structured
decisions \cite{vllm2026structured}, including through
compatible generative models. For \texttt{noul} questions,
the interface returns a probability normalized over the
\texttt{yes} and \texttt{no} token scores. This makes it
suitable for evaluating the stability of binary decisions
in LoRA-based guardrail applications.

We evaluate vLLM's structured-decision interface
using Qwen3-0.6B with a rank-16 LoRA adapter ($\alpha=32$)
fine-tuned on Civil Comments. Binary toxicity labels are derived
using a threshold of 0.5 on the dataset's toxicity scores.
After removing empty comments and exact duplicates,
we select 20,000 training, 2,000 validation, and 1,000
test examples through fixed-seed, class-proportional
sampling. The adapter is trained for three epochs,
selecting the checkpoint with the lowest validation loss.

Each request asks the \texttt{noul} question,
``Is this comment toxic?'' The interface computes a
toxicity probability by normalizing the next-token
log-probabilities of \texttt{yes} and \texttt{no}.
A probability of at least 0.5 is classified as toxic.

For each reduction scheme, we evaluate the same 1,000
comments sequentially ten times within one engine,
with prefix caching disabled. Comparing repetitions
1-9 against the first yields 9,000 comparisons per
scheme. We measure probability variation using the
fraction of changed probabilities and their mean and
maximum absolute differences. Decision stability is
evaluated through label flips, correctness transitions,
and classification accuracy.

\section{Results and Discussion}
\label{sec:results}

\subsection{Numerical Variability in LoRA Shrink}
\label{sec:shrink_results}

We examine whether numerical variation originates
within the LoRA shrink operation itself. Using identical
captured activations and adapter weights, we compare the
repeatability of Atomic Split-K=8, Split-K=1, and
Deterministic Split-K=8 across all three models, both
datasets, and LoRA ranks 16 and 32.

Across 27,440 captured shrink operations covering prefill
and the first decode step, each implementation is executed
five times, yielding 109,760 repeat comparisons per
implementation. Atomic Split-K=8 produces different FP32
outputs in 107,136 comparisons (97.61\%). In contrast,
Split-K=1 and Deterministic Split-K=8 exhibit no observed
differences across the same captured operations. 
This distinction is consistent with prior work showing
that fixed-shape GPU computations can be repeatable
when their reduction order is stable~\citep{gond2026llm}.
Our results identify unordered accumulation
in LoRA shrink as a source of run-to-run variation
under the tested execution conditions.

\begin{figure}[!htbp]
    \centering
    \includegraphics[width=\columnwidth]{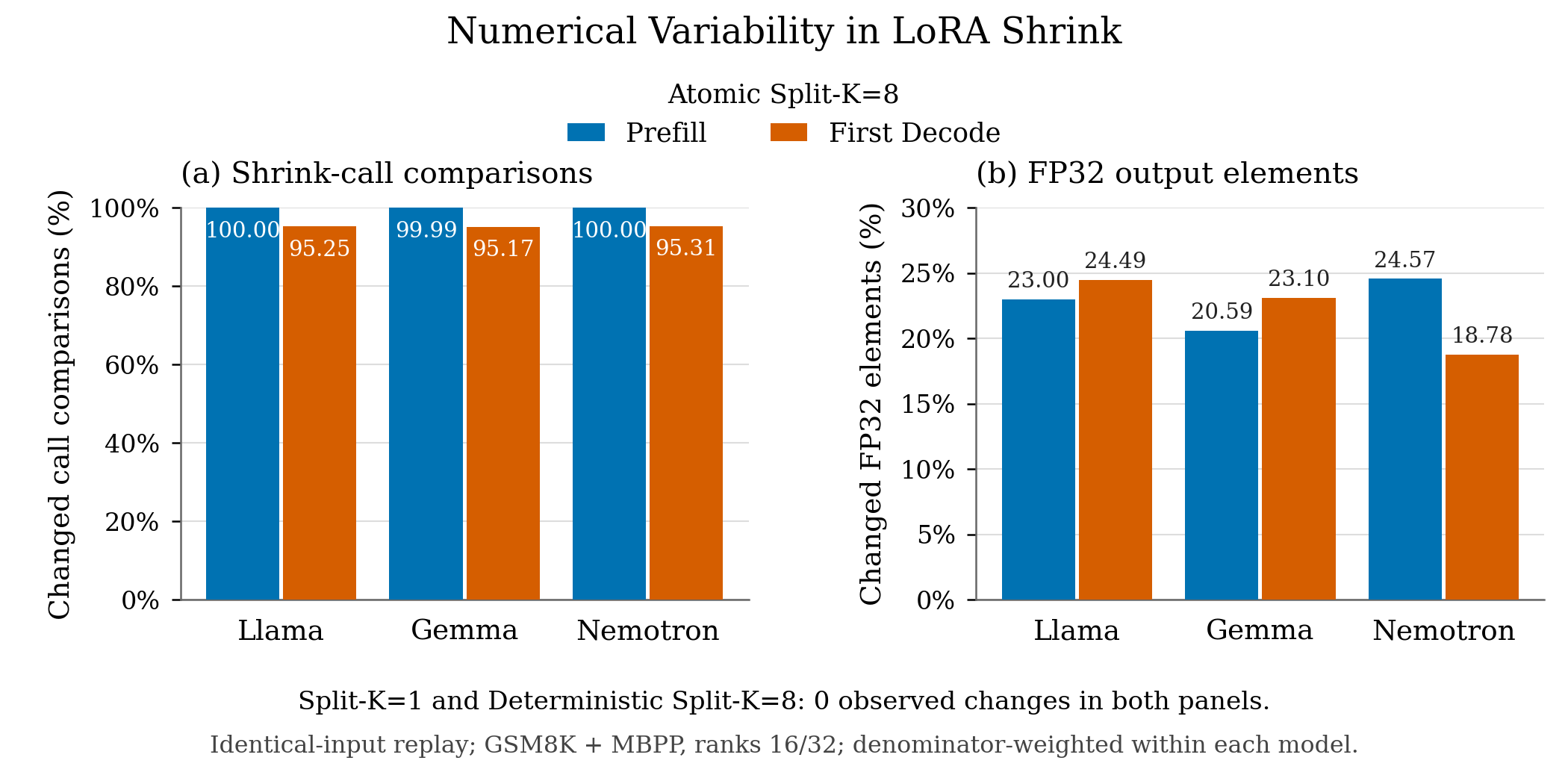}
    \caption{Run-to-run numerical variability in the LoRA
    shrink operation during prefill and first decode.
    Atomic Split-K=8 exhibits numerical variation under
    identical inputs, while Split-K=1 and Deterministic
    Split-K=8 show no observed differences.}
    \label{fig:shrink_variability}
\end{figure}

The variability occurs in both inference phases. For Atomic
Split-K=8, 99.996\% of prefill comparisons and 95.22\% of
first-decode comparisons contain at least one changed
output element (Figure~\ref{fig:shrink_variability}).
Across both phases, 22.31\% of compared FP32 elements
differ, although the mean absolute difference is only
$1.08\times10^{-8}$.

Fixed-order accumulation eliminates the observed
run-to-run variation within the tested shrink operations
without removing Split-K parallelism. To assess
reproducibility across GPU architectures, we additionally
evaluated the implementation on NVIDIA H100 and B200
GPUs under the same experimental configuration.
Neither GPU exhibited numerical variation across
repeated identical-input executions, consistent with the findings obtained on the primary NVIDIA RTX PRO 6000 Blackwell platform.

\subsection{Tracing LoRA Numerical Variability to Token Decisions}
\label{sec:divergence_trace}

To trace how LoRA shrink variability propagates through
autoregressive inference, we use Llama-3.2-3B-Instruct
with a rank-16 GSM8K adapter. Two Atomic Split-K=8
executions of the same prompt share their first 35
generated tokens but select different tokens at position
36. Both executions terminate naturally, producing final
answers of 7 (correct) and 8 (incorrect). Repeated
Split-K=1 executions produce identical captured states
and outputs.

To identify the origin of this divergence, we trace
numerical differences from prefill through the subsequent
decode steps. The first observed numerical difference originating from
the atomic LoRA shrink reduction that propagates beyond
the LoRA projection occurs in the layer-2 MLP
up-projection, at prompt
position 47. Two FP32 shrink values, 0.010345458984375 and
0.010345457121729851, differ by only
$1.86\times10^{-9}$. The first value lies exactly
at a BF16 rounding midpoint, while the second lies
slightly below it. Consequently, they convert to
different BF16 values ~\citep{srivastava2024optimistic}, separated by
$6.10\times10^{-5}$.

\begin{figure*}[t]
    \centering
    \includegraphics[width=0.85\textwidth]
    {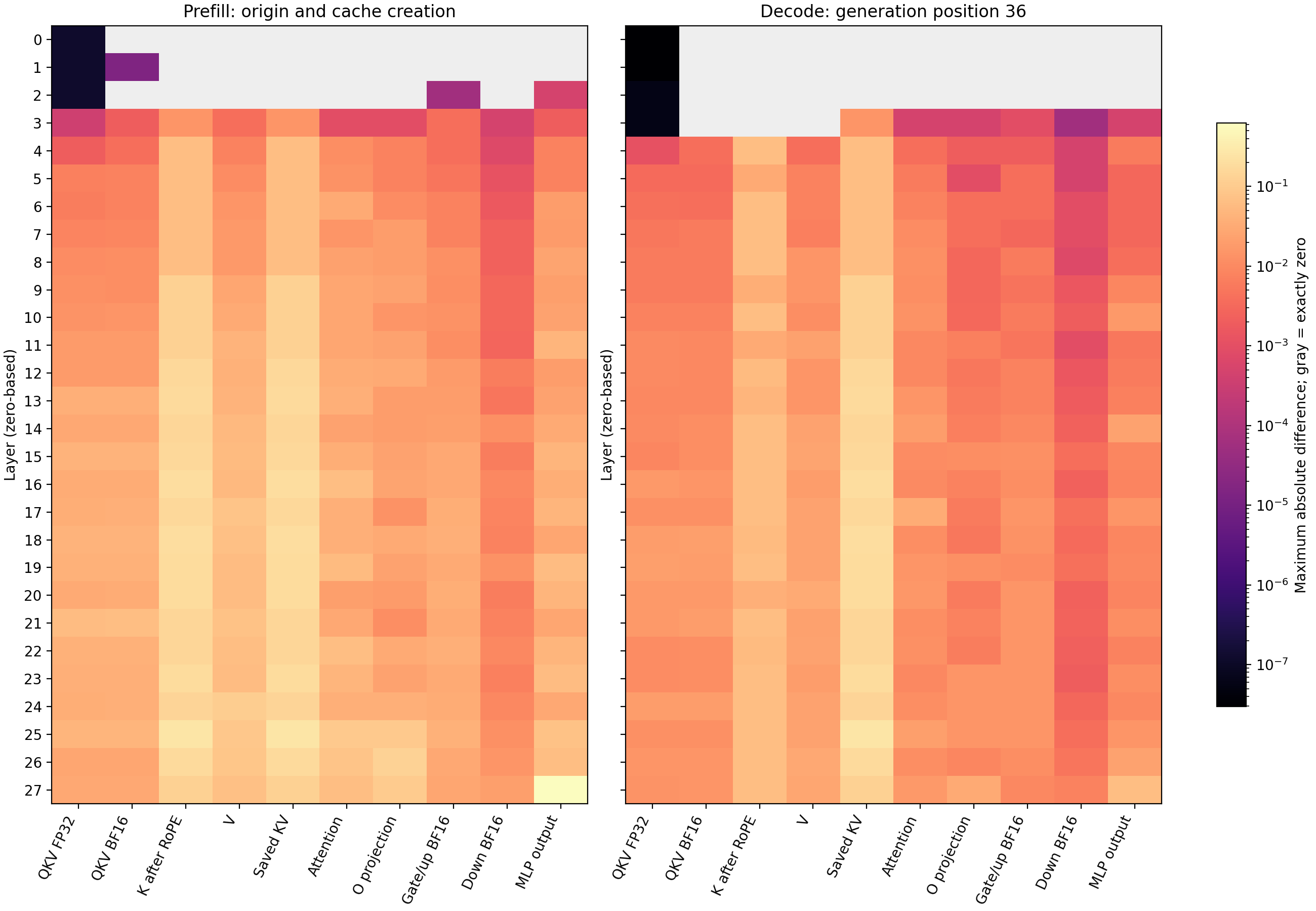}
    \caption{Numerical differences between two Atomic Split-K=8
executions during prefill and decode at generated
token 36. Colors indicate maximum absolute tensor
differences on a logarithmic scale; gray denotes
exact agreement. The first propagating difference
appears in layer 2, with cached K/V differences
from layer 3 onward. Subsequent differences reflect
both accumulated state variation and numerical
variation during later operations.}
    \label{fig:prefill_decode_trace}
\end{figure*}

This differing BF16 element affects the
subsequent MLP computation and the input to layer 3.
The resulting layer-3 KV cache contains 108 differing
key elements and 176 differing value elements at
prompt position 47. The original differing KV-cache entries persist across the subsequent 35 token generation steps.
Figure~\ref{fig:prefill_decode_trace}
illustrates the numerical differences observed
during prefill and at the divergent decode step.

To verify how shrink-output differences propagate,
we inject two observed shrink outputs into otherwise
identical executions, using Split-K=1 for all remaining
operations. This reproduces the corresponding layer-3
K/V differences at the affected prompt position,
including 284 differing cache elements. However,
the isolated shrink perturbation does not reproduce
the later token flip.

A second controlled experiment tests whether the
accumulated KV-cache differences are sufficient
to change the token decision. Immediately before generating token 36,
we substitute the incoming KV cache from either
atomic execution and continue with \mbox{Split-K=1}.
Both conditions are repeatable across two trials.
With the first cache, the logits of \texttt{There}
and \texttt{Each} are both 21.75, and the model
selects \texttt{There}, producing the correct
answer 7. With the second cache, the logit of
\texttt{There} decreases to 21.625 while
\texttt{Each} remains at 21.75. The model
therefore selects \texttt{Each} and produces
the incorrect answer 8.

The different tokens are followed by distinct
reasoning trajectories. In the correct execution,
the model subtracts the three dropped marshmallows
before dividing the remainder between the two
children:

\begin{quote}
\small
\textbf{Execution A (correct):}

There are 35 - 18 = 17 marshmallows left in the bag.

There are 17 - 3 = 14 marshmallows left in the bag.

Each kid can make 14/2 = 7 S'mores.

\textbf{Final answer: 7}
\end{quote}

In the incorrect execution, the model omits
the subtraction for the dropped marshmallows:

\begin{quote}
\small
\textbf{Execution B (incorrect):}

Each kid can have 17/2 = 8.5 S'mores.

Since you can't have half a S'more, each
kid can have 8 S'mores.

\textbf{Final answer: 8}
\end{quote}

Although the cache intervention reproduces the token
flip, it does not establish whether an individual
LoRA shrink perturbation is sufficient to cause it.
We therefore perform source-level interventions on
ten freshly generated divergent GSM8K execution
pairs using Llama-3.2-3B-Instruct with a rank-16
LoRA adapter.

We first reconstruct each atomic execution by
replaying its captured shrink outputs under otherwise
Split-K=1 execution. The resulting KV-cache states
and target-step vocabulary logits match the original
executions exactly through the first token divergence
in all ten pairs. We then identify 26 shrink operations
with identical incoming operands but differing FP32
outputs that survive BF16 conversion. Holding the
preceding execution history fixed, we substitute
individual shrink outputs with their observed
alternatives and recompute all subsequent operations
using Split-K=1. These interventions reproduce the
corresponding atomic token choices in two of the
ten prompts.

\begin{figure*}[t]
\centering
\resizebox{0.96\textwidth}{!}{%
\begin{tikzpicture}[
  x=1cm,y=1cm,
  font=\footnotesize\rmfamily,
  axis/.style={
    draw=black!72,
    line width=0.55pt
  },
  guide/.style={
    draw=black!42,
    dashed,
    line width=0.47pt
  },
  rule/.style={
    draw=black!25,
    line width=0.45pt
  },
  bluebox/.style={
    draw=black!60,
    fill=blue!10,
    line width=0.50pt,
    inner sep=3pt
  },
  orangebox/.style={
    draw=black!60,
    fill=orange!16,
    line width=0.50pt,
    inner sep=3pt
  },
  num/.style={font=\scriptsize}
]

\path[use as bounding box]
  (0,0) rectangle (16.0,4.2);


\node[anchor=west] at (0.15,3.92)
  {FP32--BF16 rounding boundary};

\node[num] at (3.55,3.49)
  {$|\Delta h_{\mathrm{FP32}}|
    =1.19\times10^{-7}$};

\draw[axis] (0.55,2.55) -- (7.10,2.55);

\draw[axis] (3.61,2.30) -- (3.61,2.86);
\draw[guide] (3.61,1.23) -- (3.61,2.25);

\fill[orange!75!black]
  (1.85,2.55) circle (2.8pt);

\fill[blue!65!black]
  (5.39,2.55) circle (2.8pt);

\node[num,anchor=south] at (1.85,2.72)
  {$h_B=-0.3857422471$};

\node[num,anchor=south] at (5.39,2.72)
  {$h_A=-0.3857421279$};

\draw[axis,->]
  (1.85,2.46) -- (1.85,1.56);

\draw[axis,->]
  (5.39,2.46) -- (5.39,1.56);

\node[orangebox] at (1.85,1.24)
  {$-0.38671875$};

\node[bluebox] at (5.39,1.24)
  {$-0.384765625$};

\node[num,anchor=north] at (3.61,0.85)
  {$-0.3857421875\quad\text{(midpoint)}$};

\node[num,text=black!63] at (3.61,0.35)
  {$|\Delta h_{\mathrm{BF16}}|
    =1.953125\times10^{-3}$};


\draw[rule] (7.55,0.35) -- (7.55,4.05);

\node[anchor=west] at (7.83,3.92)
  {Token 36:
   $m=z_{\mathrm{There}}-z_{\mathrm{Each}}$};

\draw[axis] (8.89,1.38) -- (12.27,1.38);

\draw[guide]
  (11.72,1.47) -- (11.72,3.19);

\draw[guide]
  (9.10,1.47) -- (9.10,2.00);

\draw[axis]
  (9.10,1.33) -- (9.10,1.44);

\draw[axis]
  (11.72,1.33) -- (11.72,1.44);

\node[num,anchor=north] at (9.10,1.24)
  {$-0.125$};

\node[num,anchor=north] at (11.72,1.24)
  {$0$};

\node[anchor=east] at (8.65,2.84) {A};
\node[anchor=east] at (8.65,2.03) {B};

\fill[blue!65!black]
  (11.72,2.84) circle (3pt);

\fill[orange!75!black]
  (9.10,2.03) circle (3pt);

\node[num,anchor=south] at (11.72,2.96)
  {$0$ (tie)};

\node[bluebox,anchor=west] at (12.62,2.84)
  {\texttt{There} $\to$ 7 (correct)};

\node[orangebox,anchor=west] at (12.62,2.03)
  {\texttt{Each} $\to$ 8 (incorrect)};

\end{tikzpicture}%
}

\caption{
Single-coordinate causal intervention for GSM8K
prompt 813. An observed FP32 difference crosses
a BF16 rounding boundary and changes the token
decision and final answer under controlled
Split-K=1 continuation. FP32 positions are schematic.
}

\label{fig:rounding_causal_mechanism}
\end{figure*}
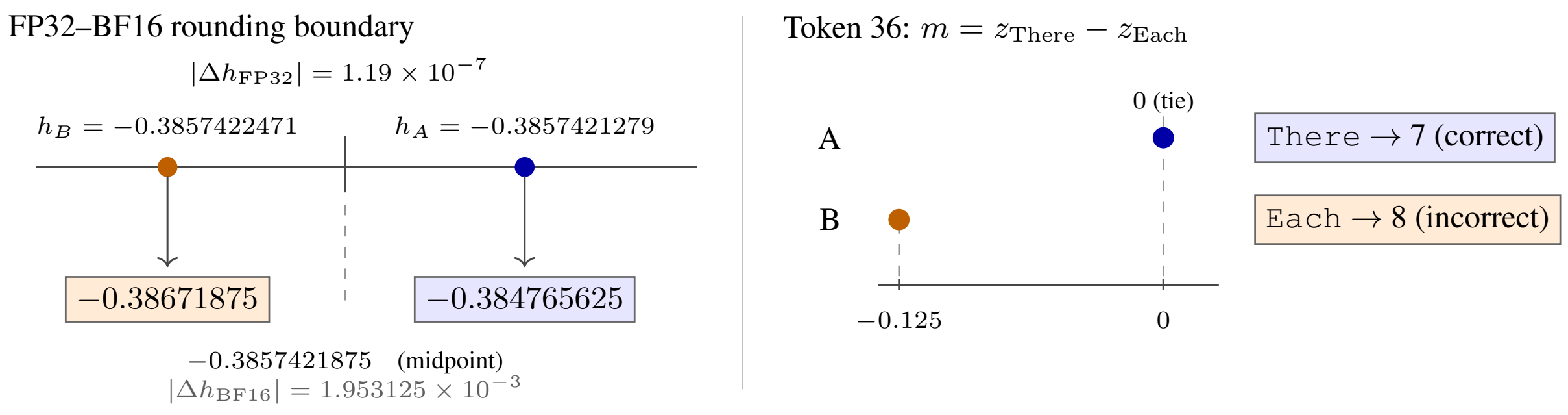

For GSM8K prompt 813, we identify a single influential
FP32 coordinate in the layer-8 QKV shrink operation
during prefill. Its observed values,
$-0.3857421278953552$ and $-0.3857422471046448$,
differ by only $1.19\times10^{-7}$ but straddle
the BF16 rounding midpoint $-0.3857421875$,
yielding $-0.384765625$ and $-0.38671875$
after conversion. Changing only this coordinate
alters 32 BF16 elements of the value projection,
which are stored in the V cache, while the
source-layer K cache remains unchanged.
Under an identical preceding execution history
and Split-K=1 for all subsequent operations,
this perturbation changes the selected token at
generation position 36 from \texttt{There} to
\texttt{Each}. Without forcing subsequent tokens,
the final answer changes from 7 (correct) to
8 (incorrect), with both continuations reproducing
exactly across repeated trials. Figure~\ref{fig:rounding_causal_mechanism}
illustrates the BF16 rounding boundary and the
resulting change in token selection.

These interventions establish that an individual
LoRA shrink perturbation can be sufficient to
alter a downstream token decision and final task
outcome. Among the ten selected divergent pairs,
we identify such an influential local source in
two cases, with one changing final-answer
correctness. Because the pairs were selected for
observed divergence, these results demonstrate
causal sufficiency in specific executions rather
than the frequency of such effects across the
benchmark.

\subsection{End-to-End Generation and Task Stability}
\label{sec:generation_results}

End-to-end generation experiments assess whether numerical variability affects generated token sequences and task correctness. Prior work has shown that numerical differences
during LLM inference can lead to divergent generations
and changes in task outcomes~\citep{yuan2025understanding}. For each model,
dataset, and LoRA rank, we perform ten repeated greedy
generations on the same 50 test prompts. Each subsequent
generation is compared with the first, yielding 2,700
comparisons per dataset and reduction method across the
six model-rank configurations.

Atomic Split-K=8 produces different token sequences in
467 of 2,700 comparisons (17.30\%) on GSM8K and
404 of 2,700 comparisons (14.96\%) on MBPP.
These differences occur across all six model-rank
configurations for both datasets. In contrast, Split-K=1
and Deterministic Split-K=8 show no observed output
differences across the corresponding repeated executions.
The atomic implementation also exhibits differences in
selected-token logits at 14.57\% of comparable
matching-prefix positions on GSM8K and 18.30\% on MBPP.

Not all output differences affect task correctness.
On GSM8K, Atomic Split-K=8 produces 151 comparisons
with different extracted final answers, of which 84
change correctness. These comprise 45 transitions from
correct to incorrect and 39 in the opposite direction.
On MBPP, 53 comparisons change whether the generated
program passes the public tests provided in the prompt.

The relationship between output variability and
correctness is not uniform across configurations.
For example, Llama on GSM8K at rank 16 exhibits
105 changed-output comparisons and 18 correctness
flips, whereas Nemotron at the same rank exhibits
77 changed outputs but 21 correctness flips.
On MBPP, Nemotron at rank 16 produces 36 changed-output
comparisons without any change in test outcomes.
Thus, the frequency of output variation alone does
not determine its effect on task correctness.

\begin{table}[!htbp]
\vspace{-2pt}
\centering
\caption{Output and correctness changes under repeated
greedy inference. Each reduction method has 2,700
comparisons per dataset, using repetitions 1-9 against
repetition 0.}
\vspace{4pt}
\label{tab:generation_stability}
\small
\begin{tabular}{llrr}
\toprule
Dataset & Reduction & Output changes & Correctness flips \\
\midrule
GSM8K & Atomic SK8 & 467 (17.30\%) & 84 \\
      & SK1        & 0 & 0 \\
      & Det. SK8   & 0 & 0 \\
\midrule
MBPP  & Atomic SK8 & 404 (14.96\%) & 53 \\
      & SK1        & 0 & 0 \\
      & Det. SK8   & 0 & 0 \\
\bottomrule
\end{tabular}
\end{table}

The deterministic implementations eliminate the observed
within-configuration output variability but do not
necessarily produce identical answers across reduction
methods or improve task accuracy. Their benefit in
these experiments is repeatability, rather than a
systematic improvement in correctness.

\subsection{Jev-Style Structured Decision Stability}
\label{sec:decision_results}

Atomic Split-K=8 produces different returned probabilities
in 7,332 of 9,000 comparisons (81.47\%). The mean absolute
probability difference is 0.00143, with a maximum of
0.05672. Despite this frequent probability variation,
only seven comparisons (0.078\%) change the binary
classification, affecting two distinct comments.
Split-K=1 and Deterministic Split-K=8 exhibit no observed
probability or decision changes across the same
repeated evaluations.

\begin{table}[!htbp]
\centering
\caption{Probability and decision stability on Civil
Comments. Each reduction method has 9,000 comparisons
across 1,000 comments.}
\vspace{4pt}
\label{tab:decision_stability}
\small
\begin{tabular}{lrrr}
\toprule
Metric & Atomic SK8 & SK1 & Det. SK8 \\
\midrule
Probability changes (\%) & 81.47 & 0 & 0 \\
Mean absolute drift       & 0.00143 & 0 & 0 \\
Maximum absolute drift    & 0.05672 & 0 & 0 \\
Decision flips            & 7 & 0 & 0 \\
Affected comments         & 2 & 0 & 0 \\
Accuracy (\%)             & 94.25 & 94.20 & 94.30 \\
\bottomrule
\end{tabular}
\end{table}

All seven decision flips occur in two comments whose
initial probability is exactly $0.5$. In the flipped
repetitions, the probability decreases to approximately
$0.46879$, changing the predicted class under the
$p\geq0.5$ threshold. Thus, frequent numerical variation
in returned probabilities does not necessarily change
the binary decision. The similar classification
accuracies across reduction methods provide no evidence
of a systematic accuracy advantage.

\subsection{Repeatability Under Different Batch Sizes}
\label{sec:batch_stability}

Batch-size experiments test whether repeatability persists when multiple requests are processed together. Using Llama-3.2-3B-Instruct
with GSM8K adapters of ranks 16 and 32, we evaluate
batch sizes of 1, 16, 32, and 64. For each configuration,
the same 64 prompts are processed in a fixed order,
with 150 tokens generated per request. We repeat each
fixed-batch workload ten times and compare repetitions
1-9 against the first.

Atomic Split-K=8 exhibits output variation at every
tested batch size. Across both adapter ranks, the
fraction of changed outputs ranges from 36.55\% at
batch size 1 to 39.93\% at batch size 32. In contrast,
Split-K=1 and Deterministic Split-K=8 produce identical
token sequences across all repetitions at each fixed
batch size.

\begin{table}[!htbp]
\centering
\caption{Output differences under fixed and changing
batch sizes, aggregated across LoRA ranks 16 and 32.}
\vspace{4pt}
\label{tab:batch_stability}
\small
\begin{tabular}{lcc}
\toprule
Reduction & Fixed batch & Cross-batch \\
\midrule
Atomic SK8 & 1748/4608 (37.93\%) & 168/384 (43.75\%) \\
SK1        & 0/4608 (0\%)       & 152/384 (39.58\%) \\
Det. SK8   & 0/4608 (0\%)       & 150/384 (39.06\%) \\
\bottomrule
\end{tabular}
\end{table}

We further compare the first execution at batch sizes
16, 32, and 64 against the first execution at batch
size 1 for the same prompts. Despite their repeatability
within each fixed batch size, Split-K=1 and
Deterministic Split-K=8 exhibit output differences
in 39.58\% and 39.06\% of these cross-batch comparisons,
respectively. Atomic Split-K=8 differs in 43.75\%,
although this comparison also includes its inherent
run-to-run variability.

Repeatability within a fixed batch size does not imply batch invariance.
Eliminating schedule-dependent accumulation in the LoRA
shrink operation is sufficient to obtain repeatable
outputs under the tested fixed-batch conditions, but
does not guarantee identical outputs across different
batch sizes. Changes in batch size may alter numerical
execution elsewhere in the inference stack, beyond the
reduction controlled by our kernel.

\subsection{Performance Cost of Deterministic Reduction}
\label{sec:performance_results}

We evaluate the performance cost of deterministic LoRA
shrink reduction by comparing Atomic Split-K=8,
Split-K=1, and Deterministic Split-K=8 under identical
inference workloads. For each model, dataset, and LoRA
rank, we measure 50 sequential requests generating
exactly 150 tokens each, across five independent engine
trials. All measurements use CUDA graph-enabled inference,
with model initialization and warmup excluded.

Across the 12 model-dataset-rank configurations,
Deterministic Split-K=8 achieves a 16.8-21.1\%
latency-based speedup over Split-K=1, averaging 19.2\%.
Compared with Atomic Split-K=8, the deterministic
implementation incurs 1.02-2.62\% higher end-to-end
latency, averaging 1.69\%. Table~\ref{tab:performance_comparison}
summarizes these results across models and adapter ranks.

\begin{table}[H]
\centering
\caption{End-to-end latency comparison of Deterministic
Split-K=8 against SK1 and Atomic Split-K=8,
averaged across GSM8K and MBPP.}
\vspace{4pt}
\label{tab:performance_comparison}
\small
\begin{tabular}{lrrr}
\toprule
Model & Rank &
\shortstack{Speedup vs.\\SK1 (\%)} &
\shortstack{Latency overhead\\vs. Atomic (\%)} \\
\midrule
Llama    & 16 & 21.0 & 1.62 \\
         & 32 & 19.5 & 1.50 \\
Gemma    & 16 & 20.6 & 2.49 \\
         & 32 & 18.9 & 2.17 \\
Nemotron & 16 & 18.3 & 1.13 \\
         & 32 & 16.8 & 1.20 \\
\midrule
Average  & -- & 19.2 & 1.69 \\
\bottomrule
\end{tabular}
\end{table}

\begin{figure}[!htbp]
    \centering
    \includegraphics[width=\columnwidth]
    {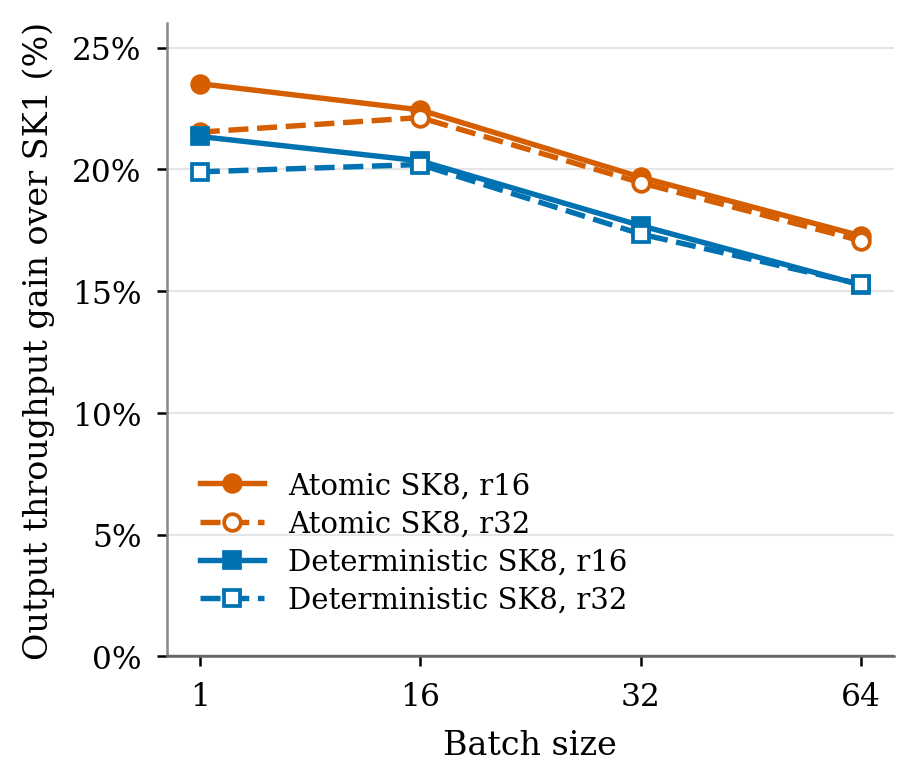}
    \caption{Output-throughput improvement of Atomic
    Split-K=8 and Deterministic Split-K=8 relative
    to Split-K=1 across batch sizes 1, 16, 32,
    and 64 using Llama on GSM8K.}
    \label{fig:batch_performance}
\end{figure}

Fixed-batch performance is measured using Llama-3.2-3B-Instruct on GSM8K at batch sizes of 1, 16, 32, and 64. Each configuration
processes the same 64 prompts across three independent
engine trials. Deterministic Split-K=8 consistently
achieves higher output throughput than Split-K=1
across both adapter ranks and all tested batch sizes,
as shown in Figure~\ref{fig:batch_performance}.

At batch size 64, Deterministic Split-K=8 achieves
6,079 and 6,060 generated tokens per second for
ranks 16 and 32, respectively, compared with
5,274 and 5,257 tokens per second for Split-K=1.
This represents approximately 15.3\% higher
throughput for both ranks. Relative to Atomic
Split-K=8, the deterministic implementation incurs
only 1.5-1.7\% lower throughput at this batch size.

The throughput advantage over Split-K=1 varies
with batch size, reaching approximately 20-21\%
at batch size 1 and remaining above 15\% at batch
size 64. Thus, deterministic reduction preserves
a substantial performance advantage over unsplit
execution even as batching increases.

From an engineering perspective, deterministic
Split-K reduction provides a targeted alternative
to disabling Split-K parallelism. The implementation
achieves repeatable inference under the tested fixed
configurations while retaining much of the performance
benefit of parallel reduction. However, the observed
cross-batch differences show that kernel-level
determinism does not guarantee system-wide
reproducibility. Evaluating numerical stability therefore requires
testing both the LoRA shrink kernel in isolation
and the complete inference pipeline under the
intended batching conditions.

\section{Limitations}
\label{sec:limitations}

Our deterministic Split-K implementation addresses
nondeterminism from unordered floating-point accumulation
in LoRA shrink but does not guarantee reproducibility
across inference configurations. Although Split-K=1 and
Deterministic Split-K=8 produce repeatable outputs at
fixed batch sizes, changing the batch size can still
alter outputs by changing execution shapes and numerical
behavior in other operations. Fixed-size or bucketed
batching and controlled prefix-cache reuse may help,
but their effectiveness and performance trade-offs
under varying workloads remain to be studied.

\section{Conclusion}
\label{sec:conclusion}

We demonstrate that atomic LoRA Split-K reduction
introduces numerical variability that can affect
model outputs and correctness. Atomic Split-K=8
changes generated outputs in 17.30\% of GSM8K
and 14.96\% of MBPP comparisons. On Civil Comments,
81.47\% of comparisons exhibit probability variation,
including seven binary decision flips.

Motivated by a numerical reproducibility problem in an internal LoRA inference
system, we developed and deployed a deterministic
Split-K=8 kernel that preserves parallelism while
combining FP32 partials in a fixed order. It eliminates
the observed within-configuration variability, achieving
an average 19.2\% speedup over Split-K=1 with 1.69\%
latency overhead relative to Atomic Split-K=8.
The results demonstrate that targeted kernel changes
can improve inference repeatability without requiring
unsplit reduction.

\bibliographystyle{mlsys2025}
\bibliography{references}
\section*{AI Usage Disclosure}
GPT-5.6 Sol was used to assist with code generation, code
review, LaTeX formatting, and grammar
revisions. Algorithm design, experimental planning, execution, data analysis,
interpretation, and scientific conclusions were carried
out by the authors.
\end{document}